\pdfoutput=1
\documentclass[11pt]{article}

\usepackage[preprint]{acl}

\usepackage{times}
\usepackage{latexsym}

\usepackage[T1]{fontenc}
\usepackage[utf8]{inputenc}

\usepackage{microtype}

\usepackage{inconsolata}

\usepackage{graphicx}

\usepackage{listings}
\title{Zero, Finite, and Infinite Belief History of Theory of Mind Reasoning in Large Language Models}

\author{Weizhi Tang \\
  University of Edinburgh\\
  \texttt{Weizhi.Tang@ed.ac.uk} \\\And
  Vaishak Belle \\
  University of Edinburgh\\
  \texttt{vbelle@ed.ac.uk} \\}

\begin{document}
\maketitle
\begin{abstract}
Large Language Models (LLMs) have recently shown a promise and emergence of Theory of Mind (ToM) ability and even outperform humans in certain ToM tasks. To evaluate and extend the boundaries of the ToM reasoning ability of LLMs, we propose a novel concept, taxonomy, and framework, the ToM reasoning with Zero, Finite, and Infinite Belief History and develop a multi-round text-based game, called \textit{Pick the Right Stuff}, as a benchmark. We have evaluated six LLMs with this game and found their performance on Zero Belief History is consistently better than on Finite Belief History. In addition, we have found two of the models with small parameter sizes outperform all the evaluated models with large parameter sizes. We expect this work to pave the way for future ToM benchmark development and also for the promotion and development of more complex AI agents or systems which are required to be equipped with more complex ToM reasoning ability\footnote{Our source code of the study and benchmark is open source and available at \url{https://anonymous.4open.science/r/Pick-the-Right-Stuff-B1F0}.}.
\end{abstract}

\section{Introduction}

Theory of Mind (ToM) is the ability to understand, attribute, and infer the mental states of oneself and others to be able to represent how others perceive, interpret, reason, and respond to the surrounding world~\citep{Baron-Cohen_Leslie_Frith_1985,Premack_Woodruff_1978,Quesque_Rossetti_2020}. In order for individuals to understand and reason about others' beliefs, intention, desires, emotions, obligations, and so on, the ability of ToM is essential~\citep{Premack_Woodruff_1978,Frith_Frith_2006,Quesque_Rossetti_2020}. And with the emergence and consistent advancement of Large Language Models (LLMs), they have demonstrated impressive and near-human-level capabilities in various tasks~\citep{Naveed_Khan_Qiu_Saqib_Anwar_Usman_Akhtar_Barnes_Mian_2024,Wan_Hu_Zhang_Wang_Wen_Lu_2024,Wang_Li_Yin_Wu_Liu_2023,Chang_Wang_Wang_Wu_Yang_Zhu_Chen_Yi_Wang_Wang_et_al._2023} and some studies have shown a promise and even an emergence of ToM reasoning ability in LLMs~\citep{Strachan_Albergo_Borghini_Pansardi_Scaliti_Gupta_Saxena_Rufo_Panzeri_Manzi_et_al._2024,Kosinski_2023,Jamali_Williams_Cai,Moghaddam_Honey}. Furthermore, since the essence of ToM is understanding and representing the mental states of others, for LLMs within its frequently mentioned and used situation, the question and answering (QA), correctly understanding the mental states, such as the intention of the users, is necessary and the key to efficiently generate appropriate, meaningful, and beneficial responses. Additionally, not only in the QA situation, LLMs are mentioned and used in various other scenarios, such as using LLMs as embodied agents~\citep{Zhang_Du_Shan_Zhou_Du_Tenenbaum_Shu_Gan_2024,Liu_Yu_Zhang_Xu_Lei_Lai_Gu_Ding_Men_Yang_et_al._2023,Gallotta_Todd_Zammit_Earle_Liapis_Togelius_Yannakakis_2024,Wang_Xie_Jiang_Mandlekar_Xiao_Zhu_Fan_Anandkumar_2023}, understanding the beliefs of other agents in environments is essential to act in a beneficial, safe, and helpful way for both themselves and other agents. Therefore, further exploration of the ToM ability of LLMs is necessary and meaningful.

In order to understand the ToM ability of LLMs, various benchmarks have been developed in different ways to handle different evaluation perspectives of the ToM ability of LLMs in previous work, such as that MindGames leverages dynamic epistemic logic to generate complex ToM problems including the k-order ToM reasoning and different problem contexts~\citep{sileo-lernould-2023-mindgames}, Hi-ToM introduces a complex k-order ToM problems based on Sally-Anne-like stories, which further contain communications between agents~\citep{wu-etal-2023-hi}, and FANToM introduces the ToM benchmark in a manner of interactions of agents and aligns the ToM problems more with real-world scenarios requiring ToM reasoning~\citep{kim-etal-2023-fantom}.

In this study, in order to further understand the ToM ability of LLMs, besides providing a text-based and multi-round game as a new benchmark to evaluate the ToM reasoning ability of LLMs in a complex and dynamic way, we have extended the current boundaries of ToM ability evaluation range by introducing a novel concept, taxonomy, and framework of ToM evaluation, which is the ToM reasoning with Zero, Finite, and Infinite Belief History. This concept and taxonomy aim to provide a clear idea and structure for evaluating a specific aspect of ToM, the beliefs, in a more advanced and nuanced manner. 

We give a brief explanation of the concept here and discuss it in more detail in Section~\ref{sec:ZFI_ToM}. The Zero Belief History means that the test subject can identify the latest beliefs of others without needing to reason from the available or known belief history while the Finite Belief History means that the test subject should and have to utilize and reason with the known finite belief history to distinguish and identify the latest beliefs of others. In addition, since recent ToM benchmarks introduce more complex ToM problems which are even not comfortable for humans, such as the high k-order ToM problems mentioned by~\citet{sileo-lernould-2023-mindgames} and~\citet{wu-etal-2023-hi}, and because ~\citet{Strachan_Albergo_Borghini_Pansardi_Scaliti_Gupta_Saxena_Rufo_Panzeri_Manzi_et_al._2024} shows LLMs, especially GPT-4~\citep{OpenAI_Achiam_Adler_Agarwal_Ahmad_Akkaya_Aleman_Almeida_Altenschmidt_Altman_et_al._2024}, sometimes can even outperform humans in ToM problems, we propose to not make the ToM evaluation of LLMs be constrained in human-level but to extend the boundaries to more complex ToM problems so that we may have confidence to put LLMs in situations which need more complex ToM reasoning. Thus, while we consider Zero and Finite Belief History to be reasonably solvable for humans, in addition to that, we further introduce the Infinite Belief History in which the test subject should maintain an infinite belief history, typically in the way of implicit patterns, and propose to discover and study it in future work.

In this study, we explain Zero, Finite, and Infinite Belief History in detail with examples. In addition, we have developed a game called \textit{Pick the Right Stuff} as both a proof-of-concept benchmark and example for future work and an out-of-box benchmark to evaluate the ToM ability of LLMs in the case of Zero Belief History and Finite Belief History. We have also evaluated 6 LLMs, in which there are 3 models in large parameter sizes and 3 models in small parameter sizes. Unsurprisingly, all LLMs perform better under the condition of the Zero Belief History than the Finite Belief History case. In addition, an unexpected and supervising finding is that two of LLMs with the small parameter sizes outperform all the three LLMs with large parameter sizes in both of Zero Belief History and Finite Belief History case, which may further bring up evidence and an answer to the question, whether increasing the sizes of the parameter of LLMs is necessary and effective to enhance model capabilities, at least in the context of ToM. Our overall contributions are listed as follows:
\begin{enumerate}
  \item We propose a novel concept, taxonomy, and framework for evaluating ToM ability of LLMs.
  \item Along with the concept we proposed, we have developed a multi-round text-based game as a new ToM benchmark to be both a proof-of-concept and example benchmark for future benchmark development and an out-of-box benchmark to evaluate the ToM ability of LLMs.
  \item We have shown that LLMs perform better in the case of Zero Belief History than in the case of Finite Belief History and also shown that LLMs in small parameter sizes can even outperform LLMs in large parameter sizes when tackling ToM problems.
\end{enumerate}

\section{Related Work}

\subsection{ToM in LLMs}
Although there has been debating around whether the ToM ability is really emerged in LLMs among previous work~\citep{Kosinski_2023,Jamali_Williams_Cai,Sap_LeBras_Fried_Choi_2023,trott2023large,bubeck2023sparks}, recent work has shown a hard-to-deny promise of emergence of ToM in LLMs ~\citep{Strachan_Albergo_Borghini_Pansardi_Scaliti_Gupta_Saxena_Rufo_Panzeri_Manzi_et_al._2024,Kosinski_2023,Jamali_Williams_Cai}, especially in the study of  ~\citet{Strachan_Albergo_Borghini_Pansardi_Scaliti_Gupta_Saxena_Rufo_Panzeri_Manzi_et_al._2024} which compares GPT-4~\citep{OpenAI_Achiam_Adler_Agarwal_Ahmad_Akkaya_Aleman_Almeida_Altenschmidt_Altman_et_al._2024} and Llama 2~\citep{Touvron_Martin_Stone_Albert_Almahairi_Babaei_Bashlykov_Batra_Bhargava_Bhosale_et_al._2023} with human performance and demonstrates their superior performance compared to humans in certain tasks such as false beliefs and misdirection while suggesting that GPT’s poor performance stems from a hyperconservative approach to committing to conclusions rather than from a genuine failure of inference.  In this work, instead of focusing on discussing and debating the emergence of ToM in LLMs, we propose a novel concept and taxonomy of an evaluation direction of the ToM ability of LLMs and provide a clear framework for evaluating the ToM ability of LLMs in this direction.

\subsection{ToM Benchmarks}
The benchmarks for evaluating ToM ability of LLMs are recently in active development such as MindGames~\citep{sileo-lernould-2023-mindgames}, FANToM~\citep{kim-etal-2023-fantom}, Hi-ToM~\citep{wu-etal-2023-hi}, ToMChallenges~\citep{ma-etal-2023-tomchallenges},  SymmToM~\citep{Sclar_Neubig_Bisk}, ToMi~\citep{le-etal-2019-revisiting}, and so on. Each of them is proposed and developed to enhance the complexity or diversity of the ToM evaluation. In addition, \citet{Ma_Sansom_Peng_Chai_2023} has proposed to taxonomize machine ToM into individual components such as belief, intentions, desires, emotions, percepts, and more, and also suggested treating LLMs as agents that are physically situated in environments and socially engaged in interactions with humans or intelligent agents. In this paper, we propose to further extend the evaluation of beliefs in terms of Zero, Finite, and Infinite Belief History and develop a benchmark as a game in which an LLM is a player to play the multi-round text-based game for evaluating its ToM reasoning ability.

\section{ToM Reasoning with Zero, Finite, and Infinite Belief History}
\label{sec:ZFI_ToM}

\subsection{Zero Belief History}

\begin{figure}
\includegraphics[width=1\columnwidth]{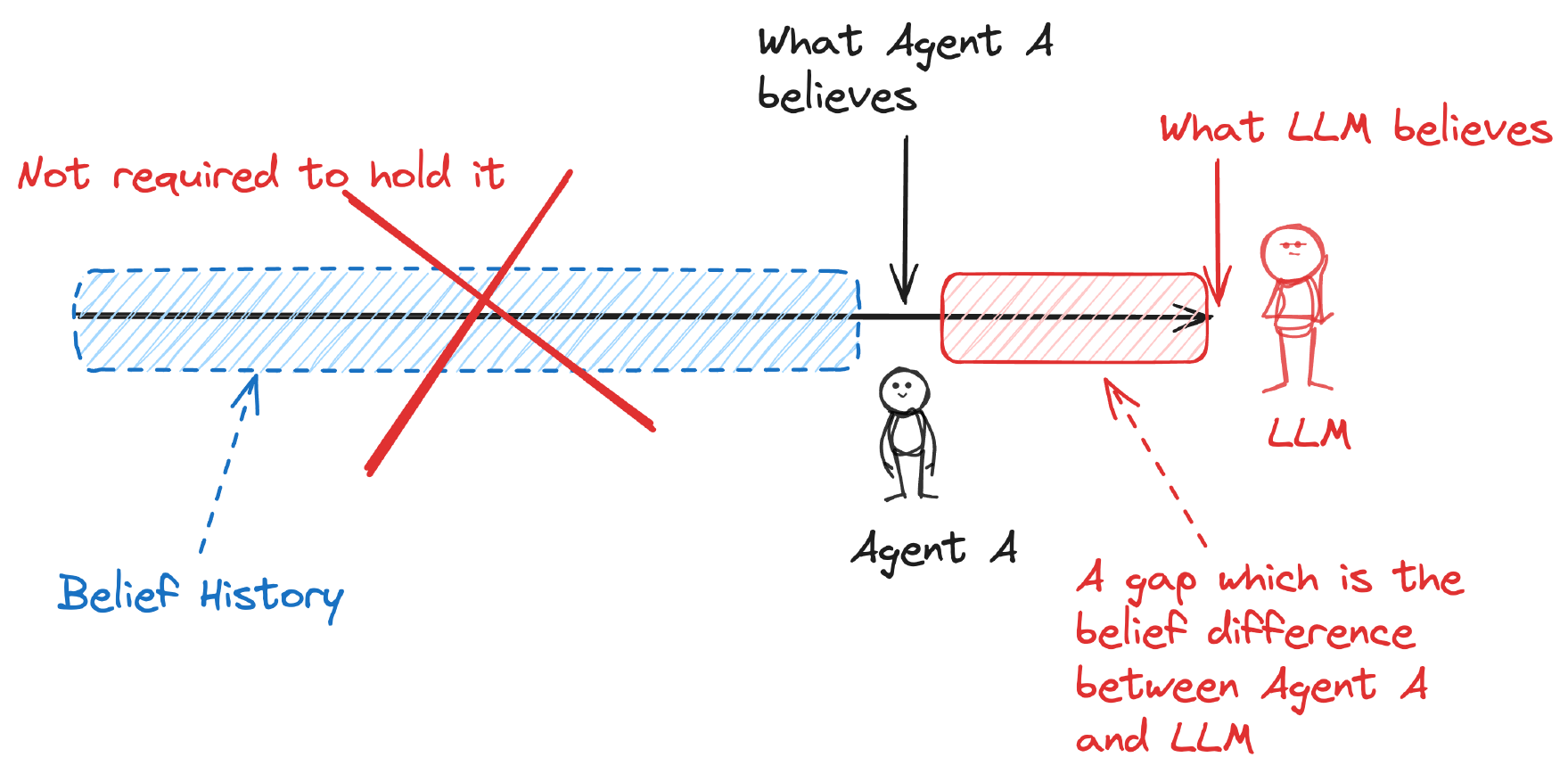}
\centering
\caption{An example of Zero Belief History in which the blue shadowed block represents the belief history which is not necessary, the red shadowed block represents a gap of beliefs between the LLM and Agent A.}  \label{fig:Zero}
\end{figure}

The ToM reasoning with Zero Belief History refers to a scenario where LLMs can directly search and find associated information from the context to perform ToM reasoning to directly distinguish and identify the beliefs of others without requiring any additional reasoning with prior knowledge and beliefs including but not limited to leveraging social rules, culture context, mathematical reasoning, and so on. For example, consider three agents discussing a certain topic in a room. If one agent leaves the room, the remaining two agents understand that the agent who left is unaware of any new content they will be discussing.

We have also demonstrated an example in Figure~\ref{fig:Zero}. The gap depicted in a red shadow block represents the belief difference between the LLM and Agent A. In order for the LLM to identify the belief of Agent A, the LLM only needs to find associated information provided in the context which is the information of the gap, without needing to use a belief history represented as a blue shadow block.

\subsection{Finite Belief History}

\begin{figure}
\includegraphics[width=1\columnwidth]{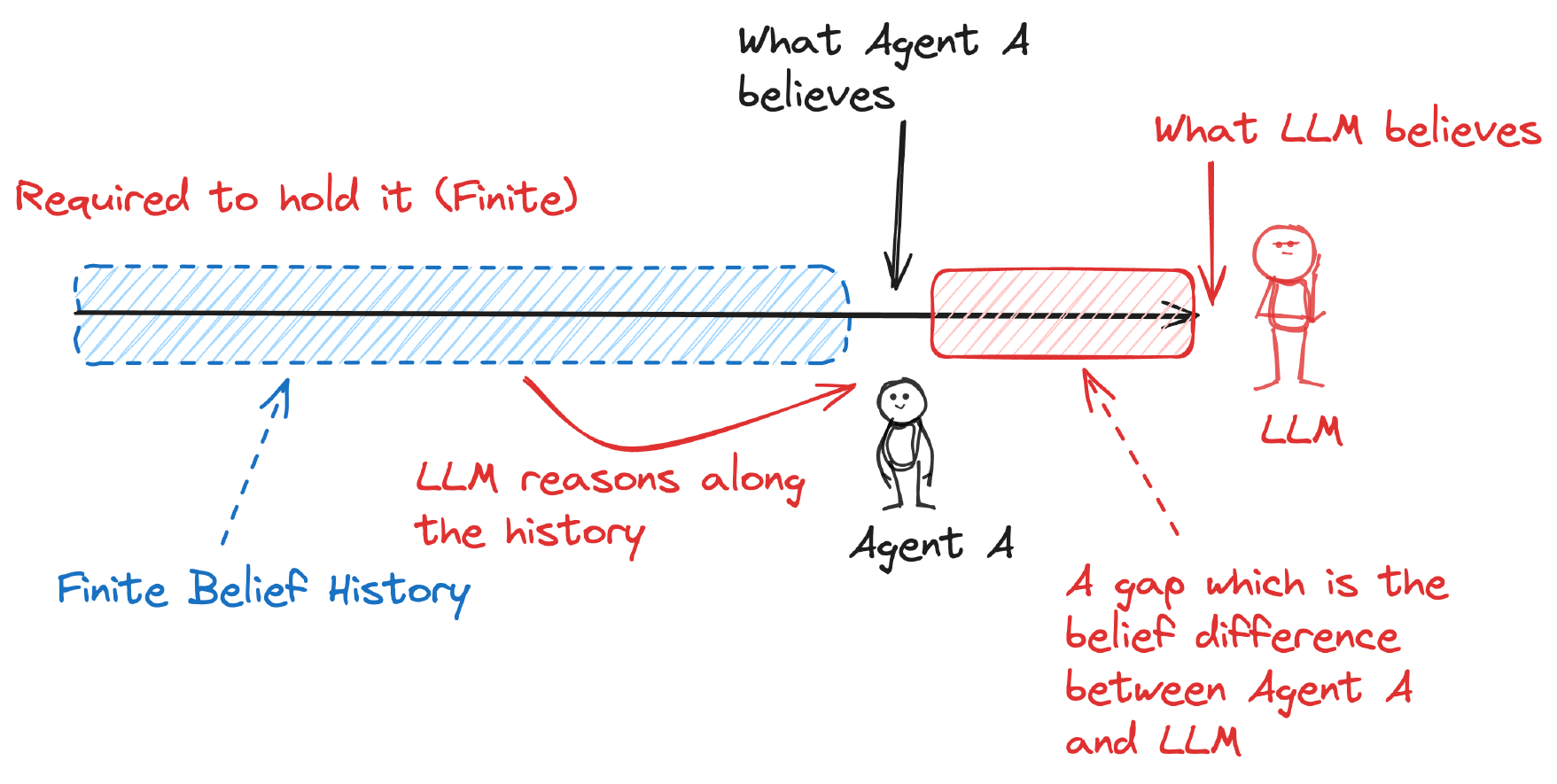}
\centering
\caption{An example of Finite Belief History in which the blue shadowed block representing a finite belief history is necessary for the LLM to reason with to identify the latest belief of the Agent A.}  \label{fig:Finite}
\end{figure}

In contrast to the ToM reasoning with the Zero Belief History, under the condition of Finite Belief History, in addition to searching and finding available and associated information from the context, due to the information is not sufficient to enable it to directly identify the latest beliefs of others, it needs and has to utilize a belief history and reason with it in a certain way, such as levering mathematical reasoning, to be able to correctly identify the latest beliefs of others. In addition, a worth noting is that, in this case, the LLM only needs to reason with a finite belief history, such as a limited number and complexity of cultural background information, social rules, or mathematical formulations.

As an example shown in Figure~\ref{fig:Finite}, besides the condition of Zero Belief History demonstrated in Figure~\ref{fig:Zero}, the LLM needs to utilize and reason with a finite belief history represented in the blue shadowed block to identify the correct latest belief of Agent A. 

\subsection{Infinite Belief History}
As a complex variant of the Finite Belief History, under the condition of Infinite Belief History, the only difference is the LLM needs and has to reason with an infinite belief history and everything else remains the same as in the Finite Belief History. An infinite belief history is typically implicitly held by the LLM in a certain pattern such as a function which represents the beliefs of the possible responses and actions towards different situations of environments in which since there could be infinite situations, responses, and actions, and the generated belief history could not be logged finitely, this function can be as a pattern to represent an infinite belief history. 

\begin{figure}
\includegraphics[width=1\columnwidth]{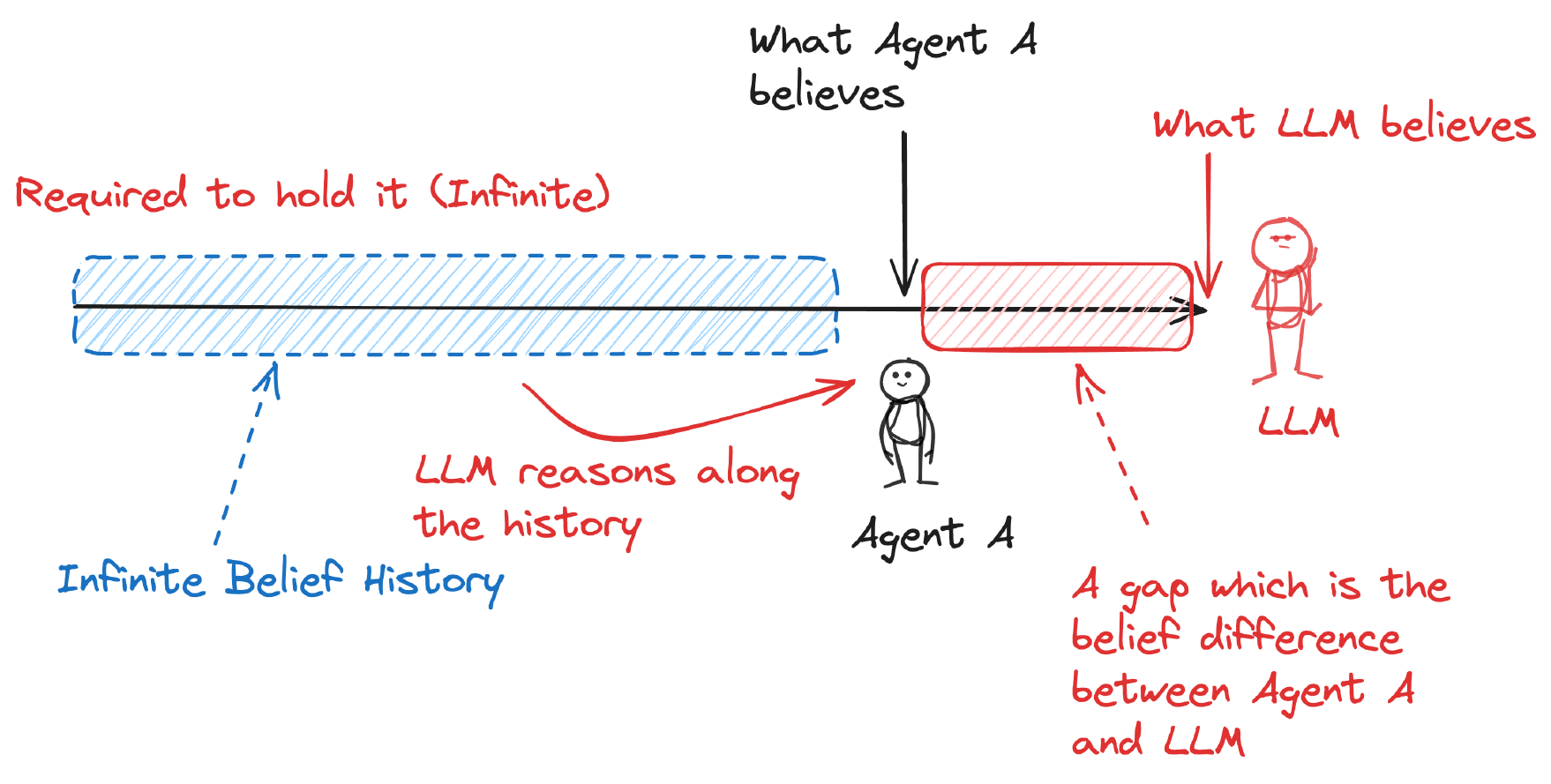}
\centering
\caption{An example of Infinite Belief History in which the LLM needs to leverage an infinite belief history represented in the blue shadowed block to reason with to identify the latest belief of the Agent A.}  \label{fig:Infinite}
\end{figure}

As depicted in Figure~\ref{fig:Infinite}, the only difference between the Finite Belief History and Infinite Belief History is that the belief history in the case of the Infinite Belief History colored in the blue shadowed block is infinite rather than finite and the LLM needs and have to utilize and reason with it to identify the latest belief of Agent A.

\section{Methodology}

\subsection{Pick the Right Stuff}
In order to evaluate the ToM reasoning ability of LLMs with each type of belief history, we develop a multi-round text-based game as a benchmark, called \textit{Pick the Right Stuff}, in which an LLM plays as a warehouse manager required to correctly pick and deliver the right stuff belonging to its original owner in the condition that the position of the stuff is not the same as its original position while the users of the warehouse do not know. This game focuses on testing ToM reasoning in the cases of Zero and Finite Belief History, while leaving the Infinite Belief History as the future work. We expect this benchmark not only as an out-of-box benchmark for ToM evaluation but also as a proof-of-concept example for future benchmark development along the concept and taxonomy of Zero, Finite, and Infinite Belief History. 

\subsubsection{Game Logic for Zero History Belief}

\begin{figure*}[t]
\includegraphics[width=1\linewidth]{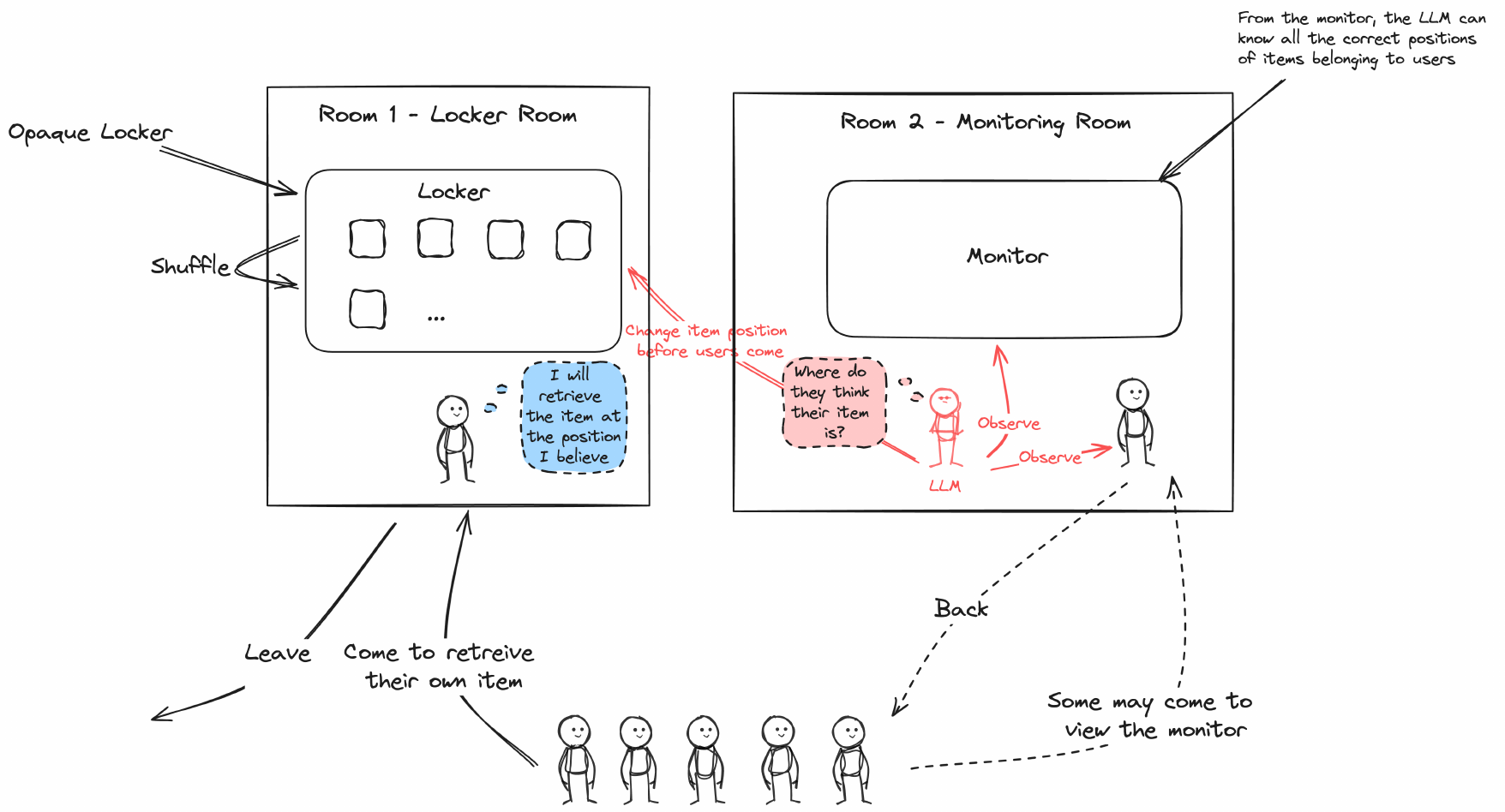}
\centering
\caption{The game logic of \textit{Pick the Right Stuff} under the condition of Zero Belief History in which there are two rooms, (1) Room 1 is the Locker Room containing an opaque locker used for storing the items of the users, and (2) Room 2 is the Monitoring Room where the LLM is located. }  \label{fig:game_under_zero}
\end{figure*}

The game, \textit{Pick the Right Stuff}, for the case of Zero Belief History, operates under the following logic\footnote{An example of the game instance under the condition of Zero Belief History is shown in Appendix~\ref{sec:appendix_zero}.}. 

There are two rooms in the game, Room 1 and Room 2. In Room 1, there is an opaque locker where the items are stored. The positions of the items in this opaque locker are randomly shuffled and reset throughout the game. Room 2 contains a monitor that allows observation of all items in the opaque locker in Room 1. An LLM is situated in this room, meaning it can observe and know the positions and owners of items at all times. 

At the onset of the game, users place their items in Room 1 and both they and the LLM are aware of the initial positions of the items. During the game, some users may randomly enter Room 2 to view the monitor and they may enter the room and view the monitor more than once times, and the LLM is also able to observe the users since in the same room. Furthermore, users may enter Room 1 at any time and expect to retrieve their items. The LLM needs to think about and provide the positions of the items as believed by the users, so that the system can automatically swap the real items with those at the given positions. If the swap is correct, the user successfully retrieves their item and the LLM scores a point. If incorrect, the LLM scores no points and the system will automatically intervene to ensure the user successfully obtains their item. After retrieval, the position of the item becomes vacant, and the retrieving user will leave. The game will continue until all positions are empty. We have also demonstrated the game logic in Figure~\ref{fig:game_under_zero}.  

In this game under the condition of Zero Belief History, the users’ beliefs are updated when they place their items at the onset of the game and when they enter Room 2 to observe the positions of the items, which means their beliefs are updated dynamically and may be changed multiple times during the game. The LLM can always know the correct positions of the items and is required to identify the false beliefs of the users in order to correctly pick and deliver the correct items to their owners to score points. Additionally, it is considered under the condition of Zero Belief History because the LLM does not need to identify the latest beliefs of the users based on a belief history but can utilize the available and associated information around the context which is whether the users enter the Room 2 to view the monitor to determine the latest beliefs of the users.

\subsubsection{Game Logic For Finite Belief History}

\begin{figure*}[t]
\includegraphics[width=1\linewidth]{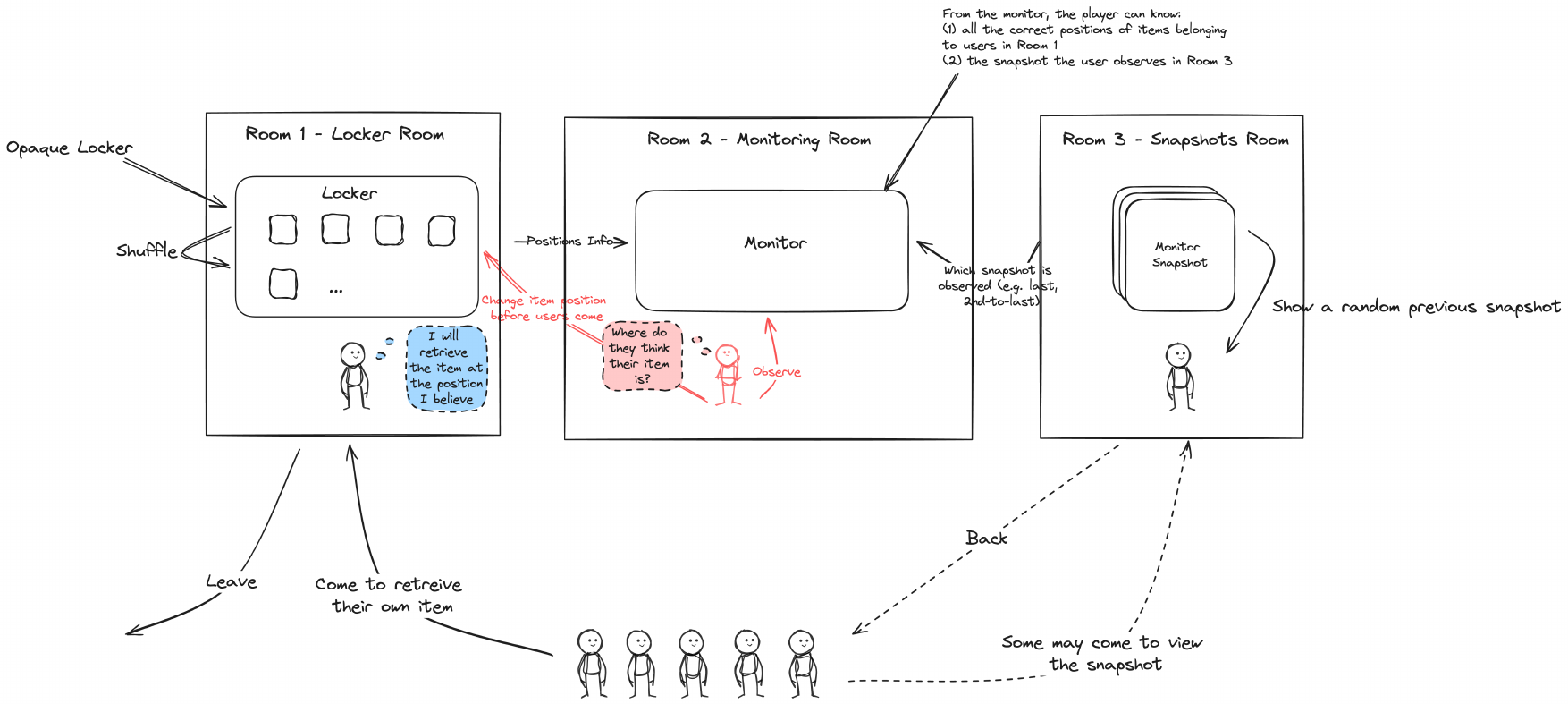}
\centering
\caption{The game logic of \textit{Pick the Right Stuff} under the condition of Finite Belief History in which there are three rooms, (1) Room 1 is the Locker Room containing an opaque locker, (2) Room 2 is the Monitoring Room where the LLM is located and enables the LLM to know the situations from both of Room 1 and Room 3, and (3) Room 3 is the Snapshots Room which shows a certain state of positions of items to users.}  \label{fig:game_under_finite}
\end{figure*}

Under the condition of Finite History Belief, the game logic is similar to that of under the condition of Zero History Belief\footnote{An example of the game instance under the condition of Finite Belief History is shown in Appendix~\ref{sec:appendix_finite}.}. The key difference lies in the addition of a third room, Room 3. In this room, there is a screen that randomly shows a snapshot representing a random previous state of the monitor located in Room 2 which is the previous state of the positions of items. Instead of entering Room 2 to observe the current monitor, some users may randomly enter Room 3 to view the snapshot on the screen during the game. In addition, the LLM located in Room 2 is only informed about the specific previous state relative to the current state in an indirect way such as "nth-to-last" instead of directly being presented. Therefore, the LLM must utilize and reason with a finite belief history to identify the latest beliefs of the users. The game logic under the condition of Finite Belief History is shown in Figure~\ref{fig:game_under_finite}.

\subsection{Experiment Settings}

In order to have a comprehensive view of the ToM reasoning ability of LLMs under Zero and Finite Belief History, we have conducted evaluations with six Large Language Models, comprising three models with large parameter sizes and three models with small parameter sizes.

For the models with large parameter sizes, we choose GPT-3.5-Turbo noted as \textit{gpt-3.5-turbo}, Llama 3 for which we select its instruct model in 70 billion parameter size noted as \textit{llama3:70b-instruct}, and Qwen~\citep{Bai_Bai_Chu_Cui_Dang_Deng_Fan_Ge_Han_Huang_et_al._2023} for which we use its chat model in 72 billion parameter size noted as \textit{qwen:72b-chat}. 
 
For the models with small parameter sizes, we select Gemma~\citep{Gemma_Team_Mesnard_Hardin_Dadashi_Bhupatiraju_Pathak_Sifre_Rivière_Kale_Love_et_al._2024} for which we use its instruct model in 7 billion parameter size noted as \textit{gemma:7b-instruct}, Mistral~\citep{Jiang_Sablayrolles_Mensch_Bamford_Chaplot_Casas_Bressand_Lengyel_Lample_Saulnier_et_al._2023} for which we choose its instruct model in 7 billion parameter size noted as \textit{mistral:7b-instruct}, and Qwen~\citep{Bai_Bai_Chu_Cui_Dang_Deng_Fan_Ge_Han_Huang_et_al._2023} for which we use its chat model in 7 billion parameter size noted as \textit{qwen:7b-chat}.
 
We then fix the temperature to 0 for all the selected LLMs to ensure the consistency of the responses. And we also fix the random seed of the game to ensure each model faces the same game states and questions. In addition, we run each model to play 60 turns of the game in the case of 5 users\footnote{Each turn consists of multiple ToM questions.}.

\section{Results and Analyses}

\begin{figure*}
\includegraphics[width=\linewidth]{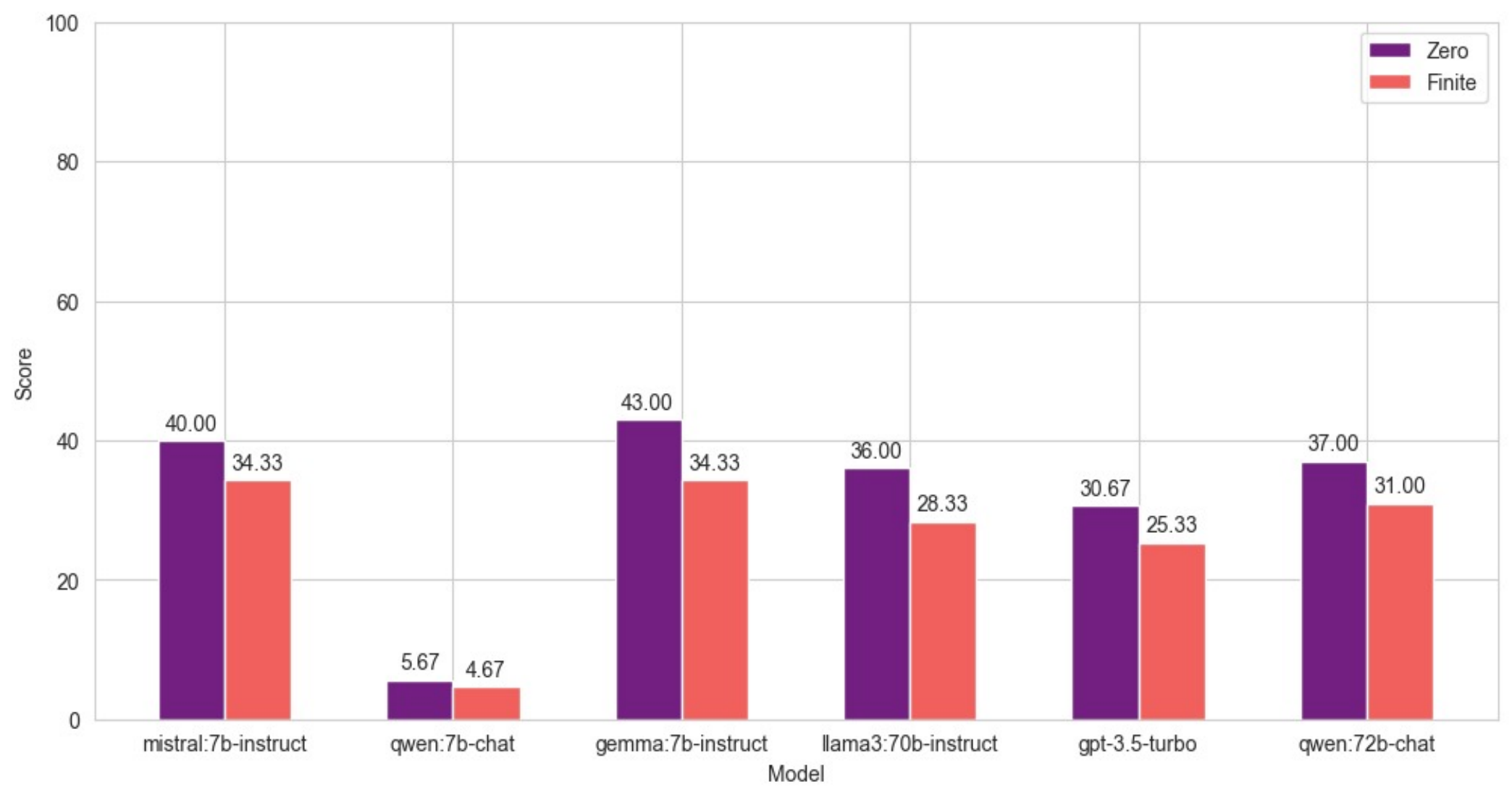}

\centering
\caption{The average scores of each model for playing the game under the condition of Zero or Finite Belief History.}  \label{fig:scores}
\end{figure*}

The average scores of the \textit{gpt-3.5-turbo}, \textit{llama3:70b-instruct}, \textit{qwen:72b-chat}, \textit{gemma:7b-instruct}, \textit{mistral:7b-instruct}, and \textit{qwen:7b-chat} playing the game, \textit{Pick the Right Stuff}, under the conditions of Zero and Finite Belief History are depicted in Figure~\ref{fig:scores}. 

For all the evaluated models, the average score under the condition of the Zero Belief History is 32.06, while it is 26.33 under the condition of the Finite Belief History, indicating that although LLMs show an impressive promise and performance on ToM reasoning, they still struggle with it. 

In addition, from the difference between the two conditions, which is 5.73, and also as shown in Figure~\ref{fig:scores}, the performance of LLMs under the condition of Finite Belief History is consistently lower than the performance under the condition of Zero Belief History, which indicates that the Finite History Belief case is indeed harder for them than the Zero History Belief.

Furthermore, among the selected LLMs, the model with small parameter size, \textit{gemma:7b-instruct}, emerged as the best-performing, achieving an average score of 43.00 under the Zero Belief History condition and 34.33 under the Finite Belief History condition, thereby outperforming all the other LLMs. In addition, \textit{mistral:7b-instruct} also shows better performance over all the other LLMs with large parameter sizes, like \textit{gpt-3.5-turbo}. Although there may be differences in the training data and architectures leveraged by these models, the outstanding performance of the models with smaller parameter sizes presents evidence and an answer to a thought-provoking question, whether increasing model parameter size may effectively enhance LLMs capabilities, at least discussed in the context of ToM reasoning ability.

%===============================================================================
\section{Implications}
We propose a novel concept, taxonomy, and framework to evaluate the ToM reasoning ability of LLMs under the conditions of Zero, Finite, and Infinite Belief History. Although we focus primarily on LLMs as the subjects of our evaluation, this concept and evaluation approach is equally applicable to other AI agents and systems. 

Additionally, since the essence and necessity of complex ToM reasoning ability in many applications such as autonomous driving, complex multi-actor story writing, embodied agents under the multi-agent case, and more, we expect this work to pave the way for more intelligent AI agents or systems which are required to equipped with more complex ToM reasoning.

%===============================================================================

\section{Conclusion}
\label{sec:conclusion}
In this study, we propose a novel concept, taxonomy, and framework to extend the boundaries of evaluation of the ToM reasoning ability of LLMs, by introducing Zero, Finite, and Infinite Belief History. In addition, we develop a multi-round text-based game as both of a proof-of-concept and example benchmark for future benchmark development along the concept proposed and an out-of-box benchmark to evaluate the ToM reasoning ability of LLMs under the conditions of Zero and Finite Belief History. We choose six LLMs of which three of them are in large parameter sizes and three of them are in small parameter sizes. We have demonstrated the evaluated LLMs perform better under the condition of Zero Belief History than under the condition of Finite Belief History. In addition, we have also found that two of the models with small parameter sizes outperform all the selected models with large parameter sizes. We hope this work can pave the way for future benchmark development and also the development of more complex AI agents and systems which are required to tackle more complex ToM problems.

%===============================================================================
\section{Limitations}
Due to this study focusing on proposing a novel evaluation concept of ToM reasoning rather than presenting an exhaustively comprehensive comparison of all available LLMs and since the unavoidable heavy computation costs incurred by GPT-4, we do not run an evaluation of GPT-4 and other LLMs. Nevertheless, future work can leverage this out-of-box game to evaluate GPT-4 and other LLMs.

In addition, the game we developed does not cover all possibilities and variants under the conditions of Zero and Finite Belief History such as replacing mathematical reasoning in the case of Finite Belief History with social rules or cultural background. We expect the future work can work along it to develop various benchmarks to evaluate ToM reasoning ability under the conditions of Zero, Finite, and Infinite Belief History with different kinds of belief history to prompt comprehensive evaluation.

Furthermore, since presenting implementation examples of Zero and Finite Belief History is sufficient to express and explain the concept and due to that the LLMs still struggle with handling ToM reasoning in the game under the conditions of the Zero and Finite Belief History, we leave the implementation of Infinite Belief History for future work.

%===============================================================================

\bibliography{anthology,custom}

\begin{thebibliography}{31}
\providecommand{\natexlab}[1]{#1}

\bibitem[{Bai et~al.(2023)Bai, Bai, Chu, Cui, Dang, Deng, Fan, Ge, Han, Huang, Hui, Ji, Li, Lin, Lin, Liu, Liu, Lu, Lu, Ma, Men, Ren, Ren, Tan, Tan, Tu, Wang, Wang, Wang, Wu, Xu, Xu, Yang, Yang, Yang, Yang, Yao, Yu, Yuan, Yuan, Zhang, Zhang, Zhang, Zhang, Zhou, Zhou, Zhou, and Zhu}]{Bai_Bai_Chu_Cui_Dang_Deng_Fan_Ge_Han_Huang_et_al._2023}
Jinze Bai, Shuai Bai, Yunfei Chu, Zeyu Cui, Kai Dang, Xiaodong Deng, Yang Fan, Wenbin Ge, Yu~Han, Fei Huang, Binyuan Hui, Luo Ji, Mei Li, Junyang Lin, Runji Lin, Dayiheng Liu, Gao Liu, Chengqiang Lu, Keming Lu, Jianxin Ma, Rui Men, Xingzhang Ren, Xuancheng Ren, Chuanqi Tan, Sinan Tan, Jianhong Tu, Peng Wang, Shijie Wang, Wei Wang, Shengguang Wu, Benfeng Xu, Jin Xu, An~Yang, Hao Yang, Jian Yang, Shusheng Yang, Yang Yao, Bowen Yu, Hongyi Yuan, Zheng Yuan, Jianwei Zhang, Xingxuan Zhang, Yichang Zhang, Zhenru Zhang, Chang Zhou, Jingren Zhou, Xiaohuan Zhou, and Tianhang Zhu. 2023.
\newblock \href {http://arxiv.org/abs/2309.16609} {Qwen technical report}.
\newblock (arXiv:2309.16609).
\newblock ArXiv:2309.16609 [cs].

\bibitem[{Baron-Cohen et~al.(1985)Baron-Cohen, Leslie, and Frith}]{Baron-Cohen_Leslie_Frith_1985}
Simon Baron-Cohen, Alan~M. Leslie, and Uta Frith. 1985.
\newblock \href {https://doi.org/10.1016/0010-0277(85)90022-8} {Does the autistic child have a “theory of mind”?}
\newblock \emph{Cognition}, 21(1):37–46.

\bibitem[{Bubeck et~al.(2023)Bubeck, Chandrasekaran, Eldan, Gehrke, Horvitz, Kamar, Lee, Lee, Li, Lundberg et~al.}]{bubeck2023sparks}
S{\'e}bastien Bubeck, Varun Chandrasekaran, Ronen Eldan, Johannes Gehrke, Eric Horvitz, Ece Kamar, Peter Lee, Yin~Tat Lee, Yuanzhi Li, Scott Lundberg, et~al. 2023.
\newblock Sparks of artificial general intelligence: Early experiments with gpt-4.
\newblock \emph{arXiv preprint arXiv:2303.12712}.

\bibitem[{Chang et~al.(2023)Chang, Wang, Wang, Wu, Yang, Zhu, Chen, Yi, Wang, Wang, Ye, Zhang, Chang, Yu, Yang, and Xie}]{Chang_Wang_Wang_Wu_Yang_Zhu_Chen_Yi_Wang_Wang_et_al._2023}
Yupeng Chang, Xu~Wang, Jindong Wang, Yuan Wu, Linyi Yang, Kaijie Zhu, Hao Chen, Xiaoyuan Yi, Cunxiang Wang, Yidong Wang, Wei Ye, Yue Zhang, Yi~Chang, Philip~S. Yu, Qiang Yang, and Xing Xie. 2023.
\newblock \href {http://arxiv.org/abs/2307.03109} {A survey on evaluation of large language models}.
\newblock (arXiv:2307.03109).
\newblock ArXiv:2307.03109 [cs].

\bibitem[{Frith and Frith(2006)}]{Frith_Frith_2006}
Chris~D. Frith and Uta Frith. 2006.
\newblock \href {https://doi.org/10.1016/j.neuron.2006.05.001} {The neural basis of mentalizing}.
\newblock \emph{Neuron}, 50(4):531–534.

\bibitem[{Gallotta et~al.(2024)Gallotta, Todd, Zammit, Earle, Liapis, Togelius, and Yannakakis}]{Gallotta_Todd_Zammit_Earle_Liapis_Togelius_Yannakakis_2024}
Roberto Gallotta, Graham Todd, Marvin Zammit, Sam Earle, Antonios Liapis, Julian Togelius, and Georgios~N. Yannakakis. 2024.
\newblock \href {https://doi.org/10.48550/arXiv.2402.18659} {Large language models and games: A survey and roadmap}.
\newblock (arXiv:2402.18659).
\newblock ArXiv:2402.18659 [cs].

\bibitem[{Jamali et~al.()Jamali, Williams, and Cai}]{Jamali_Williams_Cai}
Mohsen Jamali, Ziv~M Williams, and Jing Cai.
\newblock Unveiling theory of mind in large language models: A parallel to single neurons in the human brain.

\bibitem[{Jiang et~al.(2023)Jiang, Sablayrolles, Mensch, Bamford, Chaplot, Casas, Bressand, Lengyel, Lample, Saulnier, Lavaud, Lachaux, Stock, Scao, Lavril, Wang, Lacroix, and Sayed}]{Jiang_Sablayrolles_Mensch_Bamford_Chaplot_Casas_Bressand_Lengyel_Lample_Saulnier_et_al._2023}
Albert~Q. Jiang, Alexandre Sablayrolles, Arthur Mensch, Chris Bamford, Devendra~Singh Chaplot, Diego de~las Casas, Florian Bressand, Gianna Lengyel, Guillaume Lample, Lucile Saulnier, Lélio~Renard Lavaud, Marie-Anne Lachaux, Pierre Stock, Teven~Le Scao, Thibaut Lavril, Thomas Wang, Timothée Lacroix, and William~El Sayed. 2023.
\newblock \href {http://arxiv.org/abs/2310.06825} {Mistral 7b}.
\newblock (arXiv:2310.06825).
\newblock ArXiv:2310.06825 [cs].

\bibitem[{Kim et~al.(2023)Kim, Sclar, Zhou, Bras, Kim, Choi, and Sap}]{kim-etal-2023-fantom}
Hyunwoo Kim, Melanie Sclar, Xuhui Zhou, Ronan Bras, Gunhee Kim, Yejin Choi, and Maarten Sap. 2023.
\newblock \href {https://doi.org/10.18653/v1/2023.emnlp-main.890} {{FANT}o{M}: A benchmark for stress-testing machine theory of mind in interactions}.
\newblock In \emph{Proceedings of the 2023 Conference on Empirical Methods in Natural Language Processing}, pages 14397--14413, Singapore. Association for Computational Linguistics.

\bibitem[{Kosinski(2023)}]{Kosinski_2023}
Michal Kosinski. 2023.
\newblock \href {https://doi.org/10.48550/ARXIV.2302.02083} {Evaluating large language models in theory of mind tasks}.

\bibitem[{Le et~al.(2019)Le, Boureau, and Nickel}]{le-etal-2019-revisiting}
Matthew Le, Y-Lan Boureau, and Maximilian Nickel. 2019.
\newblock \href {https://doi.org/10.18653/v1/D19-1598} {Revisiting the evaluation of theory of mind through question answering}.
\newblock In \emph{Proceedings of the 2019 Conference on Empirical Methods in Natural Language Processing and the 9th International Joint Conference on Natural Language Processing (EMNLP-IJCNLP)}, pages 5872--5877, Hong Kong, China. Association for Computational Linguistics.

\bibitem[{Liu et~al.(2023)Liu, Yu, Zhang, Xu, Lei, Lai, Gu, Ding, Men, Yang, Zhang, Deng, Zeng, Du, Zhang, Shen, Zhang, Su, Sun, Huang, Dong, and Tang}]{Liu_Yu_Zhang_Xu_Lei_Lai_Gu_Ding_Men_Yang_et_al._2023}
Xiao Liu, Hao Yu, Hanchen Zhang, Yifan Xu, Xuanyu Lei, Hanyu Lai, Yu~Gu, Hangliang Ding, Kaiwen Men, Kejuan Yang, Shudan Zhang, Xiang Deng, Aohan Zeng, Zhengxiao Du, Chenhui Zhang, Sheng Shen, Tianjun Zhang, Yu~Su, Huan Sun, Minlie Huang, Yuxiao Dong, and Jie Tang. 2023.
\newblock \href {http://arxiv.org/abs/2308.03688} {Agentbench: Evaluating llms as agents}.
\newblock (arXiv:2308.03688).
\newblock ArXiv:2308.03688 [cs].

\bibitem[{Ma et~al.(2023{\natexlab{a}})Ma, Gao, and Xu}]{ma-etal-2023-tomchallenges}
Xiaomeng Ma, Lingyu Gao, and Qihui Xu. 2023{\natexlab{a}}.
\newblock \href {https://doi.org/10.18653/v1/2023.conll-1.2} {{T}o{MC}hallenges: A principle-guided dataset and diverse evaluation tasks for exploring theory of mind}.
\newblock In \emph{Proceedings of the 27th Conference on Computational Natural Language Learning (CoNLL)}, pages 15--26, Singapore. Association for Computational Linguistics.

\bibitem[{Ma et~al.(2023{\natexlab{b}})Ma, Sansom, Peng, and Chai}]{Ma_Sansom_Peng_Chai_2023}
Ziqiao Ma, Jacob Sansom, Run Peng, and Joyce Chai. 2023{\natexlab{b}}.
\newblock \href {http://arxiv.org/abs/2310.19619} {Towards a holistic landscape of situated theory of mind in large language models}.
\newblock (arXiv:2310.19619).
\newblock ArXiv:2310.19619 [cs].

\bibitem[{Moghaddam and Honey()}]{Moghaddam_Honey}
Shima~Rahimi Moghaddam and Christopher~J Honey.
\newblock Boosting theory-of-mind performance in large language models via prompting.

\bibitem[{Naveed et~al.(2024)Naveed, Khan, Qiu, Saqib, Anwar, Usman, Akhtar, Barnes, and Mian}]{Naveed_Khan_Qiu_Saqib_Anwar_Usman_Akhtar_Barnes_Mian_2024}
Humza Naveed, Asad~Ullah Khan, Shi Qiu, Muhammad Saqib, Saeed Anwar, Muhammad Usman, Naveed Akhtar, Nick Barnes, and Ajmal Mian. 2024.
\newblock \href {http://arxiv.org/abs/2307.06435} {A comprehensive overview of large language models}.
\newblock (arXiv:2307.06435).
\newblock ArXiv:2307.06435 [cs].

\bibitem[{OpenAI et~al.(2024)OpenAI, Achiam, Adler, Agarwal, Ahmad, Akkaya, Aleman, Almeida, Altenschmidt, Altman, Anadkat, Avila, Babuschkin, Balaji, Balcom, Baltescu, Bao, Bavarian, Belgum, Bello, Berdine, Bernadett-Shapiro, Berner, Bogdonoff, Boiko, Boyd, Brakman, Brockman, Brooks, Brundage, Button, Cai, Campbell, Cann, Carey, Carlson, Carmichael, Chan, Chang, Chantzis, Chen, Chen, Chen, Chen, Chen, Chess, Cho, Chu, Chung, Cummings, Currier, Dai, Decareaux, Degry, Deutsch, Deville, Dhar, Dohan, Dowling, Dunning, Ecoffet, Eleti, Eloundou, Farhi, Fedus, Felix, Fishman, Forte, Fulford, Gao, Georges, Gibson, Goel, Gogineni, Goh, Gontijo-Lopes, Gordon, Grafstein, Gray, Greene, Gross, Gu, Guo, Hallacy, Han, Harris, He, Heaton, Heidecke, Hesse, Hickey, Hickey, Hoeschele, Houghton, Hsu, Hu, Hu, Huizinga, Jain, Jain, Jang, Jiang, Jiang, Jin, Jin, Jomoto, Jonn, Jun, Kaftan, Kaiser, Kamali, Kanitscheider, Keskar, Khan, Kilpatrick, Kim, Kim, Kim, Kirchner, Kiros, Knight, Kokotajlo, Kondraciuk, Kondrich,
  Konstantinidis, Kosic, Krueger, Kuo, Lampe, Lan, Lee, Leike, Leung, Levy, Li, Lim, Lin, Lin, Litwin, Lopez, Lowe, Lue, Makanju, Malfacini, Manning, Markov, Markovski, Martin, Mayer, Mayne, McGrew, McKinney, McLeavey, McMillan, McNeil, Medina, Mehta, Menick, Metz, Mishchenko, Mishkin, Monaco, Morikawa, Mossing, Mu, Murati, Murk, Mély, Nair, Nakano, Nayak, Neelakantan, Ngo, Noh, Ouyang, O’Keefe, Pachocki, Paino, Palermo, Pantuliano, Parascandolo, Parish, Parparita, Passos, Pavlov, Peng, Perelman, Peres, Petrov, Pinto, Michael, Pokorny, Pokrass, Pong, Powell, Power, Power, Proehl, Puri, Radford, Rae, Ramesh, Raymond, Real, Rimbach, Ross, Rotsted, Roussez, Ryder, Saltarelli, Sanders, Santurkar, Sastry, Schmidt, Schnurr, Schulman, Selsam, Sheppard, Sherbakov, Shieh, Shoker, Shyam, Sidor, Sigler, Simens, Sitkin, Slama, Sohl, Sokolowsky, Song, Staudacher, Such, Summers, Sutskever, Tang, Tezak, Thompson, Tillet, Tootoonchian, Tseng, Tuggle, Turley, Tworek, Uribe, Vallone, Vijayvergiya, Voss, Wainwright, Wang,
  Wang, Wang, Ward, Wei, Weinmann, Welihinda, Welinder, Weng, Weng, Wiethoff, Willner, Winter, Wolrich, Wong, Workman, Wu, Wu, Wu, Xiao, Xu, Yoo, Yu, Yuan, Zaremba, Zellers, Zhang, Zhang, Zhao, Zheng, Zhuang, Zhuk, and Zoph}]{OpenAI_Achiam_Adler_Agarwal_Ahmad_Akkaya_Aleman_Almeida_Altenschmidt_Altman_et_al._2024}
OpenAI, Josh Achiam, Steven Adler, Sandhini Agarwal, Lama Ahmad, Ilge Akkaya, Florencia~Leoni Aleman, Diogo Almeida, Janko Altenschmidt, Sam Altman, Shyamal Anadkat, Red Avila, Igor Babuschkin, Suchir Balaji, Valerie Balcom, Paul Baltescu, Haiming Bao, Mohammad Bavarian, Jeff Belgum, Irwan Bello, Jake Berdine, Gabriel Bernadett-Shapiro, Christopher Berner, Lenny Bogdonoff, Oleg Boiko, Madelaine Boyd, Anna-Luisa Brakman, Greg Brockman, Tim Brooks, Miles Brundage, Kevin Button, Trevor Cai, Rosie Campbell, Andrew Cann, Brittany Carey, Chelsea Carlson, Rory Carmichael, Brooke Chan, Che Chang, Fotis Chantzis, Derek Chen, Sully Chen, Ruby Chen, Jason Chen, Mark Chen, Ben Chess, Chester Cho, Casey Chu, Hyung~Won Chung, Dave Cummings, Jeremiah Currier, Yunxing Dai, Cory Decareaux, Thomas Degry, Noah Deutsch, Damien Deville, Arka Dhar, David Dohan, Steve Dowling, Sheila Dunning, Adrien Ecoffet, Atty Eleti, Tyna Eloundou, David Farhi, Liam Fedus, Niko Felix, Simón~Posada Fishman, Juston Forte, Isabella Fulford, Leo
  Gao, Elie Georges, Christian Gibson, Vik Goel, Tarun Gogineni, Gabriel Goh, Rapha Gontijo-Lopes, Jonathan Gordon, Morgan Grafstein, Scott Gray, Ryan Greene, Joshua Gross, Shixiang~Shane Gu, Yufei Guo, Chris Hallacy, Jesse Han, Jeff Harris, Yuchen He, Mike Heaton, Johannes Heidecke, Chris Hesse, Alan Hickey, Wade Hickey, Peter Hoeschele, Brandon Houghton, Kenny Hsu, Shengli Hu, Xin Hu, Joost Huizinga, Shantanu Jain, Shawn Jain, Joanne Jang, Angela Jiang, Roger Jiang, Haozhun Jin, Denny Jin, Shino Jomoto, Billie Jonn, Heewoo Jun, Tomer Kaftan, Lukasz Kaiser, Ali Kamali, Ingmar Kanitscheider, Nitish~Shirish Keskar, Tabarak Khan, Logan Kilpatrick, Jong~Wook Kim, Christina Kim, Yongjik Kim, Jan~Hendrik Kirchner, Jamie Kiros, Matt Knight, Daniel Kokotajlo, Lukasz Kondraciuk, Andrew Kondrich, Aris Konstantinidis, Kyle Kosic, Gretchen Krueger, Vishal Kuo, Michael Lampe, Ikai Lan, Teddy Lee, Jan Leike, Jade Leung, Daniel Levy, Chak~Ming Li, Rachel Lim, Molly Lin, Stephanie Lin, Mateusz Litwin, Theresa Lopez, Ryan
  Lowe, Patricia Lue, Anna Makanju, Kim Malfacini, Sam Manning, Todor Markov, Yaniv Markovski, Bianca Martin, Katie Mayer, Andrew Mayne, Bob McGrew, Scott~Mayer McKinney, Christine McLeavey, Paul McMillan, Jake McNeil, David Medina, Aalok Mehta, Jacob Menick, Luke Metz, Andrey Mishchenko, Pamela Mishkin, Vinnie Monaco, Evan Morikawa, Daniel Mossing, Tong Mu, Mira Murati, Oleg Murk, David Mély, Ashvin Nair, Reiichiro Nakano, Rajeev Nayak, Arvind Neelakantan, Richard Ngo, Hyeonwoo Noh, Long Ouyang, Cullen O’Keefe, Jakub Pachocki, Alex Paino, Joe Palermo, Ashley Pantuliano, Giambattista Parascandolo, Joel Parish, Emy Parparita, Alex Passos, Mikhail Pavlov, Andrew Peng, Adam Perelman, Filipe de Avila~Belbute Peres, Michael Petrov, Henrique Ponde de~Oliveira Pinto, Michael, Pokorny, Michelle Pokrass, Vitchyr~H. Pong, Tolly Powell, Alethea Power, Boris Power, Elizabeth Proehl, Raul Puri, Alec Radford, Jack Rae, Aditya Ramesh, Cameron Raymond, Francis Real, Kendra Rimbach, Carl Ross, Bob Rotsted, Henri Roussez,
  Nick Ryder, Mario Saltarelli, Ted Sanders, Shibani Santurkar, Girish Sastry, Heather Schmidt, David Schnurr, John Schulman, Daniel Selsam, Kyla Sheppard, Toki Sherbakov, Jessica Shieh, Sarah Shoker, Pranav Shyam, Szymon Sidor, Eric Sigler, Maddie Simens, Jordan Sitkin, Katarina Slama, Ian Sohl, Benjamin Sokolowsky, Yang Song, Natalie Staudacher, Felipe~Petroski Such, Natalie Summers, Ilya Sutskever, Jie Tang, Nikolas Tezak, Madeleine~B. Thompson, Phil Tillet, Amin Tootoonchian, Elizabeth Tseng, Preston Tuggle, Nick Turley, Jerry Tworek, Juan Felipe~Cerón Uribe, Andrea Vallone, Arun Vijayvergiya, Chelsea Voss, Carroll Wainwright, Justin~Jay Wang, Alvin Wang, Ben Wang, Jonathan Ward, Jason Wei, C.~J. Weinmann, Akila Welihinda, Peter Welinder, Jiayi Weng, Lilian Weng, Matt Wiethoff, Dave Willner, Clemens Winter, Samuel Wolrich, Hannah Wong, Lauren Workman, Sherwin Wu, Jeff Wu, Michael Wu, Kai Xiao, Tao Xu, Sarah Yoo, Kevin Yu, Qiming Yuan, Wojciech Zaremba, Rowan Zellers, Chong Zhang, Marvin Zhang, Shengjia
  Zhao, Tianhao Zheng, Juntang Zhuang, William Zhuk, and Barret Zoph. 2024.
\newblock \href {http://arxiv.org/abs/2303.08774} {Gpt-4 technical report}.
\newblock (arXiv:2303.08774).
\newblock ArXiv:2303.08774 [cs].

\bibitem[{Premack and Woodruff(1978)}]{Premack_Woodruff_1978}
David Premack and Guy Woodruff. 1978.
\newblock \href {https://doi.org/10.1017/S0140525X00076512} {Does the chimpanzee have a theory of mind?}
\newblock \emph{Behavioral and Brain Sciences}, 1(4):515–526.

\bibitem[{Quesque and Rossetti(2020)}]{Quesque_Rossetti_2020}
François Quesque and Yves Rossetti. 2020.
\newblock \href {https://doi.org/10.1177/1745691619896607} {What do theory-of-mind tasks actually measure? theory and practice}.
\newblock \emph{Perspectives on Psychological Science}, 15(2):384–396.

\bibitem[{Sap et~al.(2023)Sap, LeBras, Fried, and Choi}]{Sap_LeBras_Fried_Choi_2023}
Maarten Sap, Ronan LeBras, Daniel Fried, and Yejin Choi. 2023.
\newblock \href {http://arxiv.org/abs/2210.13312} {Neural theory-of-mind? on the limits of social intelligence in large lms}.
\newblock (arXiv:2210.13312).
\newblock ArXiv:2210.13312 [cs].

\bibitem[{Sclar et~al.()Sclar, Neubig, and Bisk}]{Sclar_Neubig_Bisk}
Melanie Sclar, Graham Neubig, and Yonatan Bisk.
\newblock Symmetric machine theory of mind.

\bibitem[{Sileo and Lernould(2023)}]{sileo-lernould-2023-mindgames}
Damien Sileo and Antoine Lernould. 2023.
\newblock \href {https://doi.org/10.18653/v1/2023.findings-emnlp.303} {{M}ind{G}ames: Targeting theory of mind in large language models with dynamic epistemic modal logic}.
\newblock In \emph{Findings of the Association for Computational Linguistics: EMNLP 2023}, pages 4570--4577, Singapore. Association for Computational Linguistics.

\bibitem[{Strachan et~al.(2024)Strachan, Albergo, Borghini, Pansardi, Scaliti, Gupta, Saxena, Rufo, Panzeri, Manzi, Graziano, and Becchio}]{Strachan_Albergo_Borghini_Pansardi_Scaliti_Gupta_Saxena_Rufo_Panzeri_Manzi_et_al._2024}
James W.~A. Strachan, Dalila Albergo, Giulia Borghini, Oriana Pansardi, Eugenio Scaliti, Saurabh Gupta, Krati Saxena, Alessandro Rufo, Stefano Panzeri, Guido Manzi, Michael S.~A. Graziano, and Cristina Becchio. 2024.
\newblock \href {https://doi.org/10.1038/s41562-024-01882-z} {Testing theory of mind in large language models and humans}.
\newblock \emph{Nature Human Behaviour}.

\bibitem[{Team et~al.(2024)Team, Mesnard, Hardin, Dadashi, Bhupatiraju, Pathak, Sifre, Rivière, Kale, Love, Tafti, Hussenot, Sessa, Chowdhery, Roberts, Barua, Botev, Castro-Ros, Slone, Héliou, Tacchetti, Bulanova, Paterson, Tsai, Shahriari, Lan, Choquette-Choo, Crepy, Cer, Ippolito, Reid, Buchatskaya, Ni, Noland, Yan, Tucker, Muraru, Rozhdestvenskiy, Michalewski, Tenney, Grishchenko, Austin, Keeling, Labanowski, Lespiau, Stanway, Brennan, Chen, Ferret, Chiu, Mao-Jones, Lee, Yu, Millican, Sjoesund, Lee, Dixon, Reid, Mikuła, Wirth, Sharman, Chinaev, Thain, Bachem, Chang, Wahltinez, Bailey, Michel, Yotov, Chaabouni, Comanescu, Jana, Anil, McIlroy, Liu, Mullins, Smith, Borgeaud, Girgin, Douglas, Pandya, Shakeri, De, Klimenko, Hennigan, Feinberg, Stokowiec, Chen, Ahmed, Gong, Warkentin, Peran, Giang, Farabet, Vinyals, Dean, Kavukcuoglu, Hassabis, Ghahramani, Eck, Barral, Pereira, Collins, Joulin, Fiedel, Senter, Andreev, and
  Kenealy}]{Gemma_Team_Mesnard_Hardin_Dadashi_Bhupatiraju_Pathak_Sifre_Rivière_Kale_Love_et_al._2024}
Gemma Team, Thomas Mesnard, Cassidy Hardin, Robert Dadashi, Surya Bhupatiraju, Shreya Pathak, Laurent Sifre, Morgane Rivière, Mihir~Sanjay Kale, Juliette Love, Pouya Tafti, Léonard Hussenot, Pier~Giuseppe Sessa, Aakanksha Chowdhery, Adam Roberts, Aditya Barua, Alex Botev, Alex Castro-Ros, Ambrose Slone, Amélie Héliou, Andrea Tacchetti, Anna Bulanova, Antonia Paterson, Beth Tsai, Bobak Shahriari, Charline~Le Lan, Christopher~A. Choquette-Choo, Clément Crepy, Daniel Cer, Daphne Ippolito, David Reid, Elena Buchatskaya, Eric Ni, Eric Noland, Geng Yan, George Tucker, George-Christian Muraru, Grigory Rozhdestvenskiy, Henryk Michalewski, Ian Tenney, Ivan Grishchenko, Jacob Austin, James Keeling, Jane Labanowski, Jean-Baptiste Lespiau, Jeff Stanway, Jenny Brennan, Jeremy Chen, Johan Ferret, Justin Chiu, Justin Mao-Jones, Katherine Lee, Kathy Yu, Katie Millican, Lars~Lowe Sjoesund, Lisa Lee, Lucas Dixon, Machel Reid, Maciej Mikuła, Mateo Wirth, Michael Sharman, Nikolai Chinaev, Nithum Thain, Olivier Bachem,
  Oscar Chang, Oscar Wahltinez, Paige Bailey, Paul Michel, Petko Yotov, Rahma Chaabouni, Ramona Comanescu, Reena Jana, Rohan Anil, Ross McIlroy, Ruibo Liu, Ryan Mullins, Samuel~L. Smith, Sebastian Borgeaud, Sertan Girgin, Sholto Douglas, Shree Pandya, Siamak Shakeri, Soham De, Ted Klimenko, Tom Hennigan, Vlad Feinberg, Wojciech Stokowiec, Yu-hui Chen, Zafarali Ahmed, Zhitao Gong, Tris Warkentin, Ludovic Peran, Minh Giang, Clément Farabet, Oriol Vinyals, Jeff Dean, Koray Kavukcuoglu, Demis Hassabis, Zoubin Ghahramani, Douglas Eck, Joelle Barral, Fernando Pereira, Eli Collins, Armand Joulin, Noah Fiedel, Evan Senter, Alek Andreev, and Kathleen Kenealy. 2024.
\newblock \href {http://arxiv.org/abs/2403.08295} {Gemma: Open models based on gemini research and technology}.
\newblock (arXiv:2403.08295).
\newblock ArXiv:2403.08295 [cs].

\bibitem[{Touvron et~al.(2023)Touvron, Martin, Stone, Albert, Almahairi, Babaei, Bashlykov, Batra, Bhargava, Bhosale, Bikel, Blecher, Ferrer, Chen, Cucurull, Esiobu, Fernandes, Fu, Fu, Fuller, Gao, Goswami, Goyal, Hartshorn, Hosseini, Hou, Inan, Kardas, Kerkez, Khabsa, Kloumann, Korenev, Koura, Lachaux, Lavril, Lee, Liskovich, Lu, Mao, Martinet, Mihaylov, Mishra, Molybog, Nie, Poulton, Reizenstein, Rungta, Saladi, Schelten, Silva, Smith, Subramanian, Tan, Tang, Taylor, Williams, Kuan, Xu, Yan, Zarov, Zhang, Fan, Kambadur, Narang, Rodriguez, Stojnic, Edunov, and Scialom}]{Touvron_Martin_Stone_Albert_Almahairi_Babaei_Bashlykov_Batra_Bhargava_Bhosale_et_al._2023}
Hugo Touvron, Louis Martin, Kevin Stone, Peter Albert, Amjad Almahairi, Yasmine Babaei, Nikolay Bashlykov, Soumya Batra, Prajjwal Bhargava, Shruti Bhosale, Dan Bikel, Lukas Blecher, Cristian~Canton Ferrer, Moya Chen, Guillem Cucurull, David Esiobu, Jude Fernandes, Jeremy Fu, Wenyin Fu, Brian Fuller, Cynthia Gao, Vedanuj Goswami, Naman Goyal, Anthony Hartshorn, Saghar Hosseini, Rui Hou, Hakan Inan, Marcin Kardas, Viktor Kerkez, Madian Khabsa, Isabel Kloumann, Artem Korenev, Punit~Singh Koura, Marie-Anne Lachaux, Thibaut Lavril, Jenya Lee, Diana Liskovich, Yinghai Lu, Yuning Mao, Xavier Martinet, Todor Mihaylov, Pushkar Mishra, Igor Molybog, Yixin Nie, Andrew Poulton, Jeremy Reizenstein, Rashi Rungta, Kalyan Saladi, Alan Schelten, Ruan Silva, Eric~Michael Smith, Ranjan Subramanian, Xiaoqing~Ellen Tan, Binh Tang, Ross Taylor, Adina Williams, Jian~Xiang Kuan, Puxin Xu, Zheng Yan, Iliyan Zarov, Yuchen Zhang, Angela Fan, Melanie Kambadur, Sharan Narang, Aurelien Rodriguez, Robert Stojnic, Sergey Edunov, and Thomas
  Scialom. 2023.
\newblock \href {http://arxiv.org/abs/2307.09288} {Llama 2: Open foundation and fine-tuned chat models}.
\newblock (arXiv:2307.09288).
\newblock ArXiv:2307.09288 [cs].

\bibitem[{Trott et~al.(2023)Trott, Jones, Chang, Michaelov, and Bergen}]{trott2023large}
Sean Trott, Cameron Jones, Tyler Chang, James Michaelov, and Benjamin Bergen. 2023.
\newblock Do large language models know what humans know?
\newblock \emph{Cognitive Science}, 47(7):e13309.

\bibitem[{Wan et~al.(2024)Wan, Hu, Zhang, Wang, Wen, and Lu}]{Wan_Hu_Zhang_Wang_Wen_Lu_2024}
Qian Wan, Siying Hu, Yu~Zhang, Piaohong Wang, Bo~Wen, and Zhicong Lu. 2024.
\newblock \href {https://doi.org/10.1145/3637361} {"it felt like having a second mind": Investigating human-ai co-creativity in prewriting with large language models}.
\newblock \emph{Proc. ACM Hum.-Comput. Interact.}, 8(CSCW1).

\bibitem[{Wang et~al.(2023{\natexlab{a}})Wang, Xie, Jiang, Mandlekar, Xiao, Zhu, Fan, and Anandkumar}]{Wang_Xie_Jiang_Mandlekar_Xiao_Zhu_Fan_Anandkumar_2023}
Guanzhi Wang, Yuqi Xie, Yunfan Jiang, Ajay Mandlekar, Chaowei Xiao, Yuke Zhu, Linxi Fan, and Anima Anandkumar. 2023{\natexlab{a}}.
\newblock \href {http://arxiv.org/abs/2305.16291} {Voyager: An open-ended embodied agent with large language models}.
\newblock (arXiv:2305.16291).
\newblock ArXiv:2305.16291 [cs].

\bibitem[{Wang et~al.(2023{\natexlab{b}})Wang, Li, Yin, Wu, and Liu}]{Wang_Li_Yin_Wu_Liu_2023}
Xuena Wang, Xueting Li, Zi~Yin, Yue Wu, and Jia Liu. 2023{\natexlab{b}}.
\newblock \href {https://doi.org/10.1177/18344909231213958} {Emotional intelligence of large language models}.
\newblock \emph{Journal of Pacific Rim Psychology}, 17:18344909231213958.

\bibitem[{Wu et~al.(2023)Wu, He, Jia, Mihalcea, Chen, and Deng}]{wu-etal-2023-hi}
Yufan Wu, Yinghui He, Yilin Jia, Rada Mihalcea, Yulong Chen, and Naihao Deng. 2023.
\newblock \href {https://doi.org/10.18653/v1/2023.findings-emnlp.717} {Hi-{T}o{M}: A benchmark for evaluating higher-order theory of mind reasoning in large language models}.
\newblock In \emph{Findings of the Association for Computational Linguistics: EMNLP 2023}, pages 10691--10706, Singapore. Association for Computational Linguistics.

\bibitem[{Zhang et~al.(2024)Zhang, Du, Shan, Zhou, Du, Tenenbaum, Shu, and Gan}]{Zhang_Du_Shan_Zhou_Du_Tenenbaum_Shu_Gan_2024}
Hongxin Zhang, Weihua Du, Jiaming Shan, Qinhong Zhou, Yilun Du, Joshua~B. Tenenbaum, Tianmin Shu, and Chuang Gan. 2024.
\newblock \href {http://arxiv.org/abs/2307.02485} {Building cooperative embodied agents modularly with large language models}.
\newblock (arXiv:2307.02485).
\newblock ArXiv:2307.02485 [cs].

\end{thebibliography}
% \bibliography{custom}

%===============================================================================
\clearpage
\appendix

\section{Appendix}
\label{sec:appendix}

\subsection{Game Instance of Zero Belief History}
\label{sec:appendix_zero}

\begin{lstlisting}[caption={Example of the \textit{Pick the Right Stuff} game under the condition of Zero Belief History.},captionpos=t,label=lst:game_example_zero]
Welcome to, Pick the Right Stuff!

In this game, you will play the role of a warehouse manager. The warehouse contains two rooms. Room 1 is used for storing items, with each item stored in a certain position inside the opaque locker. You are situated in the Room 2, which contains a monitor that allows you to see the content of the opaque locker located in the Room 1 through the camera inside the opaque locker. Due to malfunctions in the locker system, it randomly resets the positions of the items in the opaque locker from time to time. To ensure that each user retrieves their stored item correctly, when a user comes to retrieve an item, you are required to predict the position of the item the user believes (the user will always retrieve their item based on the position they last believed). You only need to tell the system which position inside the locker the user will go to retrieve their item and then the locker system will automatically swap the item at that location with the one belonging to the user. During the game, users may or may not enter the Room 2 to observe the monitor. By observing the monitor, users will update their beliefs about the position of their item.

If a user successfully retrieves their item, you score a point and the item is removed from the locker.
If a user retrieves the wrong item, the item is returned, the user contacts the system administrator to take the correct item, and you score no points.

Indeed, this is a problematic locker system, but you are hoped to be an excellent warehouse manager!

============

Game Begins!

There are 3 users. User 0 stores its item at the position 0th of the locker. User 1 stores its item at the position 1st of the locker. User 2 stores its item at the position 2nd of the locker. 

Now they leave the room.

The locker is malfunctioning and randomly resetting the positions of the items in the locker...
The locker has returned to normal.
From the monitor, you can see the content of the locker:
The position 0th stores the item of User 1.
The position 1st stores the item of User 2.
The position 2nd stores the item of User 0.


The locker is malfunctioning and randomly resetting the positions of the items in the locker...
The locker has returned to normal.
From the monitor, you can see the content of the locker:
The position 0th stores the item of User 2.
The position 1st stores the item of User 0.
The position 2nd stores the item of User 1.


User 1 is coming to Room 1 to take his/her item...

You should only answer the position of the item the user will go to retrieve their item (e.g. 0 for the 0th, 1 for the 1st, 2 for the 2nd...).
For example, if you think the user will go to position 0th to retrieve their item, you should only answer in single number '0'.
Please make your prediction:
1
Your prediction is correct! Item in the position 1st is exchanged with the correct item in the position 2nd. User 1 successfully retrieved the item from the correct position. You score a point!

Now, from the monitor, you can see the content of the locker:
The position 0th stores the item of User 2.
The position 1st box is empty.
The position 2nd stores the item of User 0.


User 0 walks into the Room 2 and leaves the room after observing the monitor.

The locker is malfunctioning and randomly resetting the positions of the items in the locker...
The locker has returned to normal.
From the monitor, you can see the content of the locker:
The position 0th stores the item of User 0.
The position 1st stores the item of User 2.
The position 2nd box is empty.


The locker is malfunctioning and randomly resetting the positions of the items in the locker...
The locker has returned to normal.
From the monitor, you can see the content of the locker:
The position 0th stores the item of User 2.
The position 1st box is empty.
The position 2nd stores the item of User 0.


The locker is malfunctioning and randomly resetting the positions of the items in the locker...
The locker has returned to normal.
From the monitor, you can see the content of the locker:
The position 0th box is empty.
The position 1st stores the item of User 0.
The position 2nd stores the item of User 2.


User 0 is coming to Room 1 to take his/her item...

You should only answer the position of the item the user will go to retrieve their item (e.g. 0 for the 0th, 1 for the 1st, 2 for the 2nd...).
For example, if you think the user will go to position 0th to retrieve their item, you should only answer in single number '0'.
Please make your prediction:
2
Your prediction is correct! Item in the position 2nd is exchanged with the correct item in the position 1st. User 0 successfully retrieved the item from the correct position. You score a point!

Now, from the monitor, you can see the content of the locker:
The position 0th box is empty.
The position 1st stores the item of User 2.
The position 2nd box is empty.


The locker is malfunctioning and randomly resetting the positions of the items in the locker...
The locker has returned to normal.
From the monitor, you can see the content of the locker:
The position 0th box is empty.
The position 1st box is empty.
The position 2nd stores the item of User 2.


The locker is malfunctioning and randomly resetting the positions of the items in the locker...
The locker has returned to normal.
From the monitor, you can see the content of the locker:
The position 0th stores the item of User 2.
The position 1st box is empty.
The position 2nd box is empty.


The locker is malfunctioning and randomly resetting the positions of the items in the locker...
The locker has returned to normal.
From the monitor, you can see the content of the locker:
The position 0th box is empty.
The position 1st box is empty.
The position 2nd stores the item of User 2.


User 2 is coming to Room 1 to take his/her item...

You should only answer the position of the item the user will go to retrieve their item (e.g. 0 for the 0th, 1 for the 1st, 2 for the 2nd...).
For example, if you think the user will go to position 0th to retrieve their item, you should only answer in single number '0'.
Please make your prediction:
2
Your prediction is correct! Item in the position 2nd is exchanged with the correct item in the position 2nd. User 2 successfully retrieved the item from the correct position. You score a point!

Now, from the monitor, you can see the content of the locker:
The position 0th box is empty.
The position 1st box is empty.
The position 2nd box is empty.


Correct: 3
Final score: 100

Game Over!
Do you want to play another turn?(Y/n)
\end{lstlisting}

\subsection{Game Instance of Finite Belief History}
\label{sec:appendix_finite}

\begin{lstlisting}[caption={Example of the \textit{Pick the Right Stuff} game under the condition of Finite Belief History.},captionpos=t,label=lst:game_example_finite]
Game client is running!
Welcome to, Pick the Right Stuff!

In this game, you will play the role of a warehouse manager. The warehouse contains three rooms. Room 1 is used for storing items, with each item stored in a certain position inside the opaque locker. You are situated in the Room 2, which contains a monitor that allows you to see the content of the opaque locker located in the Room 1 through the camera inside the opaque locker. Due to malfunctions in the locker system, it randomly resets the positions of the items in the opaque locker from time to time. To ensure that each user retrieves their stored item correctly, when a user comes to retrieve an item, you are required to predict the position of the item the user believes (the user will always retrieve their item based on the position they last believed). You only need to tell the system which position inside the locker the user will go to retrieve their item and then the locker system will automatically swap the item at that location with the one belonging to the user. Additionally, Room 3 contains a screen which will randomly show a certain previous snapshot of the monitor located in Room 2. During the game, users may or may not enter the Room 3 to observe a certain snapshot of the monitor. By observing the snapshot, users will update their beliefs about the position of their item.

If a user successfully retrieves their item, you score a point and the item is removed from the locker.
If a user retrieves the wrong item, the item is returned, the user contacts the system administrator to take the correct item, and you score no points.

Indeed, this is a problematic locker system, but you are hoped to be an excellent warehouse manager!

============

Game Begins!

There are 3 users. User 0 stores its item at the position 0th of the locker. User 1 stores its item at the position 1st of the locker. User 2 stores its item at the position 2nd of the locker. 

Now they leave the room.

The locker is malfunctioning and randomly resetting the positions of the items in the locker...
The locker has returned to normal.
From the monitor, you can see the content of the locker:
The position 0th stores the item of User 1.
The position 1st stores the item of User 2.
The position 2nd stores the item of User 0.


The locker is malfunctioning and randomly resetting the positions of the items in the locker...
The locker has returned to normal.
From the monitor, you can see the content of the locker:
The position 0th stores the item of User 2.
The position 1st stores the item of User 0.
The position 2nd stores the item of User 1.


User 1 is coming to Room 1 to take his/her item...

You should only answer the position of the item the user will go to retrieve their item (e.g. 0 for the 0th, 1 for the 1st, 2 for the 2nd...).
For example, if you think the user will go to position 0th to retrieve their item, you should only answer in single number '0'.
Please make your prediction:
1
Your prediction is correct! Item in the position 1st is exchanged with the correct item in the position 2nd. User 1 successfully retrieved the item from the correct position. You score a point!

Now, from the monitor, you can see the content of the locker:
The position 0th stores the item of User 2.
The position 1st box is empty.
The position 2nd stores the item of User 0.


User 0 walks into the Room 3 and is observing the snapshot of the monitor...

User 0 observes the snapshot which depicts the 2nd-to-last state of the monitor and leaves the room.

User 2 is coming to Room 1 to take his/her item...

You should only answer the position of the item the user will go to retrieve their item (e.g. 0 for the 0th, 1 for the 1st, 2 for the 2nd...).
For example, if you think the user will go to position 0th to retrieve their item, you should only answer in single number '0'.
Please make your prediction:
2
Your prediction is correct! Item in the position 2nd is exchanged with the correct item in the position 0th. User 2 successfully retrieved the item from the correct position. You score a point!

Now, from the monitor, you can see the content of the locker:
The position 0th stores the item of User 0.
The position 1st box is empty.
The position 2nd box is empty.


User 0 is coming to Room 1 to take his/her item...

You should only answer the position of the item the user will go to retrieve their item (e.g. 0 for the 0th, 1 for the 1st, 2 for the 2nd...).
For example, if you think the user will go to position 0th to retrieve their item, you should only answer in single number '0'.
Please make your prediction:
1
Your prediction is correct! Item in the position 1st is exchanged with the correct item in the position 0th. User 0 successfully retrieved the item from the correct position. You score a point!

Now, from the monitor, you can see the content of the locker:
The position 0th box is empty.
The position 1st box is empty.
The position 2nd box is empty.


Correct: 3
Final score: 100

Game Over!
Do you want to play another turn?(Y/n)
\end{lstlisting}

\end{document}